\pdfoutput=1

\documentclass[11pt]{article}

\usepackage[]{emnlp2021}

\usepackage{times}
\usepackage{latexsym}
\usepackage{algorithm,algorithmicx}
\usepackage{algpseudocode}
\usepackage{framed}
\usepackage{varwidth}

\usepackage{xcolor}
\colorlet{shadecolor}{blue!20}

\usepackage{microtype}
\usepackage{booktabs}
\usepackage{multirow}
\usepackage{multicol}
\usepackage{array}
\usepackage{amsmath}
\usepackage{arydshln}
\usepackage{bbm}
\usepackage{graphicx}
\usepackage{graphics}
\usepackage{siunitx}
\usepackage{amsfonts}
\usepackage{makecell}
\usepackage{algorithm}
\usepackage[T1]{fontenc}

\usepackage[utf8]{inputenc}

\usepackage{microtype}

\usepackage[autostyle, english=american]{csquotes}
\MakeOuterQuote{"} 
\NewDocumentCommand{\semih}{ mO{} }{\textcolor{orange}{\textsuperscript{\textit{Semih}}\textsf{\textbf{\small[#1]}}}}

\title{Task-adaptive Pre-training and Self-training are Complementary for Natural Language Understanding}

\author{Shiyang Li \\
  
  UC Santa Barbara \\
  \texttt{shiyangli@ucsb.edu} \\\And
  Semih Yavuz \\
  Salesforce Research \\
  \texttt{syavuz@salesforce.com} \\ \AND
    Wenhu Chen\\ 
    UC Santa Barbara \\ 
  \texttt{wenhuchen@ucsb.edu} \\ \And
  Xifeng Yan \\
  UC Santa Barbara \\
  \texttt{xyan@cs.ucsb.edu} \\
  
  }

\begin{document}
\maketitle
\begin{abstract}
Task-adaptive pre-training (TAPT) and Self-training (ST) have emerged as the major semi-supervised approaches to improve natural language understanding (NLU) tasks with massive amount of unlabeled data. However, it's unclear whether they learn similar representations or they can be effectively combined. In this paper, we show that TAPT and ST can be complementary with simple TFS protocol by following \textbf{T}APT $\rightarrow$ \textbf{F}inetuning  $\rightarrow$ \textbf{S}elf-training (TFS) process. Experimental results show that TFS protocol can effectively utilize unlabeled data to achieve strong combined gains consistently across six datasets covering sentiment classification, paraphrase identification, natural language inference, named entity recognition and dialogue slot classification. We investigate various semi-supervised settings and consistently show that gains from TAPT and ST can be strongly \textit{additive} by following TFS procedure. We hope that TFS could serve as an important semi-supervised baseline for future NLP studies.
\end{abstract}

\section{Introduction}

Deep neural networks~\citep{goodfellow2016deep} often require large amounts of labeled data to achieve state-of-the-art performance \citep{Xie_2020_CVPR}.
However, acquiring high-quality annotations is both time-consuming and cost-expensive, which inspires research on methods that can exploit unlabeled data to improve performance \citep{He2020Revisiting}. Pre-trained language models like BERT \citep{devlin2019bert}, RoBERTa \citep{liu2019roberta} and T5 \citep{raffel2020t5} can learn general language understanding abilities from large-scale unlabeled corpora and have reduced this annotation cost. In this paradigm, large neural networks are first pre-trained on massive amounts of unlabeled data in a self-supervised manner and then finetuned on large amount of labeled data for specific downstream tasks, which has led to large improvements for natural language understanding on standard benchmarks \citep{wang2018glue,Wang2019SuperGLUEAS}. However, their success still relies on large amount of data during finetuning stage. For example, \citet{wu2020todbert} shows that BERT only achieves 6.4\% joint goal accuracy with 1\% finetuning data for dialogue state tracking task, a core component of task-oriented dialogue systems, making it far behind its full counterpart 45.6\%. This data-intensive finetuning poses several challenges for many real-world applications, where collecting large amount of labeled data is not only cost-expensive and time-consuming, but also infeasible sometimes due to data access and privacy constraints \citep{wang2021adaptive}.

Semi-supervised learning \citep{ThomasSSL} provides a plausible solution to address aforementioned data hungry issue by making effective use of freely available unlabeled data. One of the most popular semi-supervised learning algorithms is self-training \citep{ScudderST}. In self-training, a teacher model is first trained on available labeled data and then used to generate pseudo labels for unlabeled data. The original hand-annotated labeled data and the pseudo-labeled data are combined together to train a student model. The student model is assigned as a teacher model in next round and the teacher-student training procedure is repeated until convergence or reaching maximum rounds. Self-training utilizes unlabeled data in a \textit{task-specific} way during pseudo labeling process  \citep{Chen2020BigSM} and has been successfully applied to a variety of tasks, including image recognition \citep{Xie_2020_CVPR,Zoph2020RethinkingPA}, automatic speech recognition \citep{Kahn2020STSpeech}, text classification \citep{Du2020SelftrainingIP,Mukherjee2020UncertaintyawareSF}, sequence labeling \citep{wang2021adaptive}
and neural machine translation \citep{He2020Revisiting}.

Recently, task-adaptive pre-training (TAPT) \citep{gururangan2020domainadaptive} was further proposed, which can adapt pre-trained language models, e.g. BERT and RoBERTa, to unlabeled in-domain training set to improve performance \citep{gururangan2020domainadaptive}. The intuition of TAPT is that datasets for specific tasks may only contain a subset of the text within the broader domain and continuing pre-training on the task dataset itself or other relevant data can be useful \citep{gururangan2020domainadaptive}. TAPT tends to adapt its linguistic representation by utilizing the unlabeled data in a \textit{task-agnostic} way  \citep{Chen2020BigSM}. With the recent success of task-adaptive pre-training and self-training in natural language understanding (NLU), a research question arises: \textit{Are task-adaptive pre-training (TAPT) and self-training (ST) complementary for natural language understanding (NLU)?}

In this paper, we show that TAPT and ST can be complementary with simple TFS protocol by following \textbf{T}APT $\rightarrow$ \textbf{F}inetuning  $\rightarrow$ \textbf{S}elf-training process (TFS). TFS protocol
follows three steps: (1) TAPT on unlabeled corpus drawn from a task (2) Standard supervised finetuning on labeled data inheriting parameters from TAPT as initialization to train a teacher model (3) Teacher model generates pseudo labels for \textit{the same} unlabeled corpus in (1) and trains a student model in a self-training framework until convergence or reaching maximum rounds as shown in Figure \ref{fig:tsf_pipeline}. The first step utilizes unlabeled corpus in a task-agnostic way to learn general linguistic representations while the third one utilizes unlabeled corpus in a task-specific way during pseudo-labeling process. Therefore, unlabeled data are utilized \textit{twice} through two different ways by taking advantages of TAPT and ST. %
TFS can effectively utilize unlabeled data to achieve strong combined gains of TAPT and ST consistently across six datasets covering sentiment classification, paraphrase identification, natural language inference, named entity recognition and dialogue slot classification. We further investigate various semi-supervised settings and consistently show that gains from TAPT and ST can be strongly \textit{additive} by following TFS procedure.

\section{Related Work}
\noindent{\textbf{Pre-training.}} Unsupervised or self-supervised pre-training have achieved remarkable successes in natural language processing \citep{devlin2019bert,liu2019roberta,radford2019gpt2,raffel2020t5,Brown2020LanguageMA}. However, these models are pre-trained on a very large general domain corpus, e.g. Wikipedia, and may limit their performance on a specific task due to distribution shift \citep{Lee2020BioBERTAP,wu2020todbert,gururangan2020domainadaptive}. To better handle aforementioned issue, domain-adaptive pre-training (DAPT) by continuing pre-training of existing language models, e.g. BERT and RoBERTa, on a large corpus of unlabeled domain-specific text data has been proposed and achieved great successes in specific domains \citep{gururangan2020domainadaptive,Lee2020BioBERTAP,wu2020todbert}. \citet{Lee2020BioBERTAP} proposed BioBERT by continuing pre-training of BERT on biomedical domain corpus and outperformed BERT in biomedical text mining significantly. Following a similar idea, \citet{wu2020todbert} proposed ToD-BERT by continuing pre-training of BERT on nine dialogue datasets for NLU tasks in task-oriented dialogue systems and achieved great successes in various few-shot NLU tasks in dialogue domain. \citet{gururangan2020domainadaptive} took one step further and continued pre-training of language models on a much smaller amount of unlabeled data but drawn from the same distribution for a given task (TAPT), which not only can achieve competitive results with DAPT but also is complementary with it.

\noindent{\textbf{Self-training.}} Self-training as one of the earliest and simplest semi-supervised learning has recently shown state-of-the-art performance for tasks like image classification \citep{Xie_2020_CVPR,Sun2019LearningTS}, object detection \citep{Zoph2020RethinkingPA} and can perform at par with fully supervised models while using much less labeled training data. On natural language processing, \citet{Mukherjee2020UncertaintyawareSF} applied self-training for few-shot text classification and incorporated uncertainty estimation of the underlying neural network for unlabeled data selection. \citet{wang2021adaptive} improved self-training with meta-learning by adaptive sample reweighting to mitigate error propagation from noisy pseudo-labels for named entity recognition and slot tagging in task-oriented dialog systems. \citet{He2020Revisiting} injected noise to the input space as a noisy version of self-training for neural sequence generation and obtained state-of-the-art performance for tasks like neural machine translation. \citet{Du2020SelftrainingIP} utilized information retrieval to retrieve task-specific in-domain data from a large amount of web sentences for self-training. Beyond these applications of self-training, \citet{wei2021theoretical} further theoretically proved that self-training and input-consistency
regularization will achieve high accuracy in regard to ground-truth labels under certain assumptions. 

There also exist works combing pre-training with self-training. \citet{Chen2020BigSM} first conducted self-supervised pre-training with SimCLR \citep{Chen2020ASF} on ImageNet \citep{Russakovsky2015ImageNetLS} in a task-agnostic way, then finetuned pre-trained models on limited labeled data and finally did self-training/knowledge distillation \citep{Hinton2015DistillingTK} via the same unlabeled examples as pre-training in a task-specific way. Such a framework enables models to make use of data twice in both pre-training and self-training/knowledge distillation stage. \citet{Xu2020SelftrainingAP} followed this framework on speech recognition and achieved state-of-the-art performance only with very limited labeled data. However, it's unclear that language models like BERT that have already been pre-trained in a very large general corpus can benefit this framework or not since \citet{Chen2020BigSM} and \citet{Xu2020SelftrainingAP} conducted pre-training from scratch. In addition, they only did self-training in one round, making it unclear whether iterative self-training without pre-training can achieve comparable results in the end. A recent work \citep{Du2020SelftrainingIP} did both continuing pre-training and self-training in retrieved data from open domains but only observe gains for self-training while our work utilizes existing in-domain unlabeled data and found that both TAPT and self-training are effective.

\section{Algorithms}
\begin{figure*}[h!]
  \centering
  \includegraphics[scale=0.50]{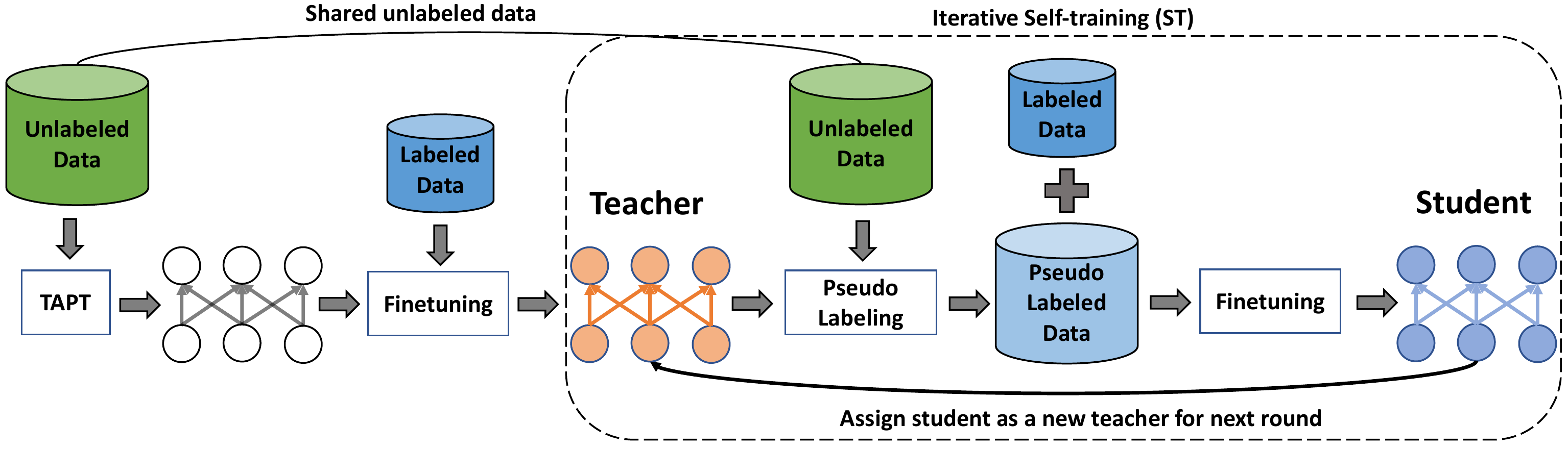}
  \caption{The overall pipeline of TFS. It has three steps (1) TAPT on unlabeled corpus drawn from a task (2) Train a teacher model on labeled data with TAPT as initialization (3) Teacher generates pseudo labels from \textit{share} unlabeled corpus with (1) and trains a student model with both labeled and pseudo labeled data in an iterative self-training framework.}
  \label{fig:tsf_pipeline}
\end{figure*}
\subsection{Problem setup}

Denote $D_{l} =\{x_{i}$, $y_{i}\}^{N}$ to be a set of $N$ labeled instances, where $x_{i}$ is a sequence of $m$ tokens: $x_{i} = \{ x_{i1}, x_{i2}, ..., x_{im} \}$  with $y_{i}$ being its label. Also, consider $D_{u} = \{x_{j}\}^{M}$ to be a set of $M$ unlabeled instances drawn from the same distribution of $\{x_{i}\}^{N}$, where $M \gg N$. Assuming that we can only access a small amount of labeled data along with a much larger amount of unlabeled data, our goal is to fully leverage unlabeled data $D_{u}$ to improve model performance.

\subsection{Task-adaptive Pre-training (TAPT)}

One simple yet effective way to improve BERT-like models with unlabeled data is task-adaptive pre-training (TAPT). The approach of TAPT is quite straightforward -- simply continuing pre-training BERT-like models with masked language modeling (MLM) \citep{devlin2019bert} on unlabeled text data for a specific given task \citep{gururangan2020domainadaptive}.

Specifically, during MLM process, a proportion of randomly sampled tokens in the input are masked out with special token \texttt{[MASK]}. We conduct dynamical token masking during pre-training following \citep{liu2019roberta,wu2020todbert}. The training objective of MLM is the cross entropy loss to reconstruct masked tokens:
\begin{equation}\label{eq:mlm_obj}
  \mathcal{L_{\text{mlm}}}  = -\sum_{j=1}^{M}\sum_{k=1}^{m} \mathbbm{1}*\text{log} (\ p(x_{jk})),
\end{equation}
where $\mathbbm{1}$ is 1 if $x_{jk}$ is masked out in the input, otherwise 0.

\subsection{Self-training (ST)} 

Self-training begins with a \textit{teacher} model $p^{t}$ trained on the labeled data $D_{l}$. The teacher model is used to generate pseudo labels for unlabeled data $D_{u}$. The augmented data $D_{l} \cup D_{u}$ is then used to train a student model $p^{s}$. Specifically, $\forall x_{j} \in D_{u}$, we use teacher model to generate its soft label and then student model is trained with standard cross-entropy loss for labeled data and KL divergence for unlabeled data, which can be formulated as:
\begin{equation}\label{eq:st_obj}
\begin{aligned}
  \mathcal{L_{\text{st}}}  = -\sum_{(x_{i},y_{i}) \in D_{l}} \text{log}  (y_{i}|p^{s}(x_{i})) \\
  - \sum_{x_{j} \in D_{u}} \text{KL} (p^{t}(x_{j})||p^{s}(x_{j}))  ,
\end{aligned}
\end{equation}
where teacher model $p^{t}$ is fixed in the current round. After training of student model with objective $\mathcal{L_{\text{st}}}$, it is assigned as a new teacher model in the next round and the teacher-student training procedure is repeated until convergence or reaching maximum rounds.

\subsection{\textbf{T}APT $\rightarrow$ \textbf{F}inetuning  $\rightarrow$ \textbf{S}elf-training (TFS)}

Although TAPT has been proven effective to utilize unlabeled data, it's \textit{task-agnostic} in the sense that it's unaware of specific tasks, e.g. sentence classification or name entity recognition. This paradigm learns general linguistic representations buried under unlabeled data, which are not directly tailored to a specific task. Utilizing data in a \textit{task-agnostic} may lose the information of unlabeled data key to the task at hand. On the contrary, self-training utilizes unlabeled data in a \textit{task-specific} way. Pseudo labels are obtained through trained models and task-specific information can be encoded into pseudo labels. However, this method may only work well when a considerable portion of the predictions on unlabeled data are correct \citep{He2020Revisiting}, otherwise early mistakes made by \textit{teacher} model $p^{t}$ due to limited labeled data can reinforce itself by generating incorrect labels for unlabeled data and re-training on this data can even result in a worse student model $p^{s}$ in the next round \citep{ssl2009Zhu}.

\begin{algorithm}[b!]

\textbf{Input}: Labeled corpus $D_{l}$, unlabeled corpus $D_{u}$ and initialized model $p_{\theta}$
\begin{algorithmic}[1]
 \State Update model $p_{\theta}$ with TAPT on unlabeled corpus $D_{u}$ by Equation \ref{eq:mlm_obj}
 
 \State Train a teacher model $p_{\tau}$ initialized with $p_{\theta}$ by finetuning on labeled corpus $D_{l}$
 
 \Repeat
 
 \State Apply $p_{\tau}$ to the unlabeled corpus $D_{u}$ to obtain $\hat{D}_{u} := \{(x_{j},p_{\tau}(x_{j}))|\forall x_{j} \in {D}_{u}\}$
 
 \State Train a student model $p_{\tau}$ on ${D}_{l} \cup \hat{D}_{u}$ by Equation \ref{eq:st_obj}
 
\State Assign $p_{\tau}$ as a teacher for the next round

\Until{Convergence or maximum rounds are reached}
\end{algorithmic}
 \caption{\label{alg:tfs_algo} TFS Protocol}
 
\end{algorithm}

TFS protocol by following \textbf{T}APT $\rightarrow$ \textbf{F}inetuning  $\rightarrow$ \textbf{S}elf-training (TFS) process can take advantages of TAPT and ST but at the same time avoid their weakness. The overall pipeline of TFS is shown in Figure \ref{fig:tsf_pipeline}.  TFS first utilizes unlabeled data in a \textit{task-agnostic} way by TAPT to have a better initialization for finetuning in next step and then finetunes a teacher model initializing its parameters from TAPT on labeled data in a standard supervised way. These two steps can build a better teacher model, avoid early mistakes and generate more accurate predictions for students, which is key to the success of self-training. The unlabeled data is leveraged again during self-training process in a \textit{task-specific} way to further boost the performance of models at hand.  We summarize the workflow of TFS in Algorithm \ref{alg:tfs_algo}.

\begin{table*} [t!]
\centering
\begin{tabular}{llccl}
\hline
\textbf{Dataset} & \textbf{Task} &  \textbf{Train size}  &  \textbf{Number of classes} & \textbf{Evaluation metrics} \\
\hline
SST-2 & Sentiment analysis & 67,349 & 2 & Accuracy \\
\hline
QNLI & Natural language inference & 104,743 & 2 & Accuracy\\
\hline
MNLI & Natural language inference & 100,000* & 3 & Accuracy\\
\hline
QQP & Paraphrase identification & 100,000* & 2 & F1\\
\hline
CoNLL 2003 & Named entity recognition & 14,041 & 9 &  F1 \\
\hline
MultiWOZ 2.1 & Slot classification & 56,557 & 30 & Micro-F1 \\
\hline
\end{tabular}
\caption{\label{statistics}
Dataset summary for evaluation. * are datasets that we randomly sample  100K instances from original training sets due to the high cost of iterative  self-training.
}
\end{table*}

\section{Experiments}
Here we conduct comprehensive experiments and analysis on different NLU datasets to demonstrate the effectiveness of TFS.

\subsection{Experimental Setup}
We use six popular large-scale datasets covering sentiment classification, paraphrase identification, natural language inference, named entity recognition and dialogue slot classification as follows. 

(1) \textbf{SST-2} consists of sentences from movie reviews and human annotations of their sentiment \citep{socher-etal-2013-recursive} . The task is to predict the sentiment of a given sentence to be positive or negative \citep{wang2018glue}. 

(2) Both \textbf{QNLI} \citep{wang2018glue} and \textbf{MNLI} \citep{williams-etal-2018-broad} are natural language inference datasets. QNLI is adapted from the SQuAD \citep{Rajpurkar2016SQuAD10} question answering dataset and the task is to predict whether the context sentence includes the answer to a given question \citep{wang2018glue}, which can be regarded as a binary classification problem. MNLI is slightly different from QNLI as it has multiple genres. Specifically, a premise sentence and a hypothesis sentence are given for each example in MNLI, and the task is to predict whether the given premise entails (\textit{entailment}), contradicts (\textit{contradiction}) the given hypothesis, or neither of them (\textit{neutral}) \citep{wang2018glue}.

(3) \textbf{QQP} is a paraphrase identification dataset \citep{Chen2018QQP}. The goal is to determine if two questions asked on Quora are semantically equivalent  \citep{wang2018glue}, which can also be formulated as a binary classification problem.

(4) \textbf{CoNLL 2003} is a name entity recognition dataset and the task is to recognize four types of named entities: \textit{persons}, \textit{locations}, \textit{organizations} and \textit{miscellaneous} entities, where \textit{miscellaneous} type does not fall into any of the previous three categories \citep{tjong-kim-sang-de-meulder-2003-introduction}.

(5) \textbf{MultiWOZ 2.1} is a large-scale multi-domain dialogue dataset with human-human conversations \citep{Eric2019MultiWOZ2M}. We convert each dialogue into turns and the task is to predict whether a slot, e.g. restaurant name, is mentioned in a turn and can be cast as a multi-label binary slot classification problem \citep{li2021coco}.

SST-2, QNLI, MNLI and QQP datasets are from GLUE benchmark \footnote{We only consider datasets with training data size larger than 10K in GLUE benchmark.} and we only report their results on development sets as extensive experiments don't allow us to submit predictions on their test sets to the official leaderboard due to submission limitations \footnote{See more about FAQ 1 at \url{https://gluebenchmark.com/faq}}. Note that for both MNLI and QQP, we randomly downsample their training sets into 100K and development sets into 5K otherwise iterative self-training in various semi-supervised setups can be too costly and for MNLI, we report results on the matched development set. On both CoNLL 2003 and MultiWOZ 2.1, we report results of their test sets. For SST-2, MNLI and QNLI, we use standard accuracy metric and for QQP and CoNLL 2003  we report their F1 scores. For MultiWOZ 2.1, we report micro-F1. We summarize details of each dataset including task, full training data size, number of classes and their evaluation metric in Table \ref{statistics}.  \\

\noindent{\textbf{TAPT.}} We use \texttt{BERT-base} and \texttt{BERT-large} as our backbone to leverage both labeled and unlabeled data. Both labeled and unlabeled data are used for TAPT in our implementation so that we can use the same checkpoint for different data split and labeled data size without repeating costly pre-training process on the same dataset. During TAPT process, we use MLM objective with random token masking probability 0.15 for each training set listed in Table \ref{statistics} following previous work \cite{devlin2019bert,liu2019roberta}.%

\begin{table*}[t!]
\centering
\begin{tabular}{cccccc}
\toprule
\textbf{Dataset} & \textbf{Model} &  \textbf{FT} &  \textbf{TAPT}  &  \textbf{ST} &  \textbf{TFS} \\
\hline
\multirow{2}{*}{SST-2} & $\texttt{BERT}_{\texttt{base}}$ & $87.3_{1.5}$ & $88.5_{0.7}$ (+1.2) & $88.4_{1.0}$ (+1.1)  & {$\textbf{89.4}_{0.8}$ (+\textbf{2.1})} \\
 & $\texttt{BERT}_{\texttt{large}}$ & $89.0_{4.2}$  & $90.7_{0.7}$ (+1.7) & $90.1_{4.2}$ (+1.1) & $\textbf{91.4}_{0.4}$ (+\textbf{2.4})  \\
\hline
\multirow{2}{*}{QNLI} & $\texttt{BERT}_{\texttt{base}}$ & $79.1_{0.8}$ & $82.0_{0.5}$ (+2.9) & $80.2_{0.7}$ (+1.1) & $\textbf{83.1}_{0.6}$ (+\textbf{4.0}) \\
 & $\texttt{BERT}_{\texttt{large}}$ & $82.6_{0.4}$ & $83.2_{0.6}$ (+0.6) & $83.7_{0.4}$ (+1.1) & $\textbf{84.4}_{0.6}$ (+\textbf{1.8}) \\
\hline

\multirow{2}{*}{MNLI} & $\texttt{BERT}_{\texttt{base}}$ & $57.3_{1.9}$ & $58.8_{1.3}$ (+1.5) & $59.2_{2.1}$ (+1.9)  & {$\textbf{60.9}_{1.4}$ (+\textbf{3.6})} \\
 & $\texttt{BERT}_{\texttt{large}}$ & $66.4_{2.6}$  & $67.6_{1.5}$ (+1.2) & $68.7_{2.2}$ (+2.3) & $\textbf{69.4}_{1.4}$ (+\textbf{3.0})  \\
\hline
\multirow{2}{*}{QQP} & $\texttt{BERT}_{\texttt{base}}$ & $71.3_{0.8}$ & $74.3_{0.8}$ (+3.0) & $72.3_{0.6}$ (+1.0) & $\textbf{75.1}_{0.9}$ (+\textbf{3.8}) \\
 & $\texttt{BERT}_{\texttt{large}}$ & $73.1_{1.7}$ & $75.1_{0.9}$ (+2.0) & $74.2_{1.8}$ (+1.1) & $\textbf{76.1}_{0.9}$ (+\textbf{3.0}) \\
\hline

\multirow{2}{*}{CoNLL 2003}  & $\texttt{BERT}_{\texttt{base}}$ & $78.8_{1.1}$ & $79.3_{1.6}$ (+0.5) & $81.6_{1.1}$ (+2.8) & $\textbf{82.2}_{1.3}$ (+\textbf{3.4}) \\
 & $\texttt{BERT}_{\texttt{large}}$ &  $76.3_{2.4}$ & $79.8_{1.0}$ (+3.5) & $79.4_{2.4}$ (+3.1) & $\textbf{82.2}_{1.1}$ (+\textbf{5.9}) \\
\hline
\multirow{2}{*}{MultiWOZ 2.1}  & $\texttt{BERT}_{\texttt{base}}$ & $75.6_{0.7}$ & $79.8_{0.5}$ (+4.2) & $76.6_{0.8}$ (+1.0) &  $\textbf{80.2}_{0.4}$ (+\textbf{4.6}) \\
 & $\texttt{BERT}_{\texttt{large}}$ & $77.7_{0.4}$ & $81.4_{0.2}$ (+3.7) & $78.7_{0.6}$ (+1.0) & $\textbf{81.8}_{0.3}$ (+\textbf{4.1}) \\

\toprule
\end{tabular}
\caption{\label{main_results}
Results comparison (\%) of finetuned baselines on labeled data (FT),  TAPT, ST and TFS of $\texttt{BERT}_{\texttt{base}}$ and $\texttt{BERT}_{\texttt{large}}$ on six different datasets with 1\% labeled data. Mean results along with their standard deviation in the subscript are listed and values inside the parentheses are gains over FT.}
\end{table*}

\noindent{\textbf{Finetuning.}} We follow standard supervised fine-tuning paradigm \citep{devlin2019bert} by adding a linear projection layer with weight $ W \in  \mathbb{R}^{K \times I}$ on top of BERT in labeled data for each dataset listed in Table \ref{statistics}, where $K$ is the number of classes and $I$ is the dimensionality of representations of BERT. Specifically, for SST-2, QNLI, MNLI and QQP, we pass the representation of \texttt{[CLS]} token $H^{CLS}$ to a linear layer followed by a \texttt{Softmax} function.  Models are trained with cross-entropy loss between the predicted distributions $\texttt{Softmax$(W(H^{CLS}))$}$ and their ground truth labels. For CoNLL 2003 name entity recognition task, we feed the representation of each token into a linear layer followed by a \texttt{Softmax} function. Models are trained with average cross-entropy loss between the predicted distributions and their labels over all tokens \footnote{We only calculate loss of the first token for words with multiple tokens after tokenization.}. For multi-label binary slot classification task on MultiWOZ 2.1, we pass the representation of \texttt{[CLS]} token $H^{CLS}$ into a linear layer followed by a \texttt{Sigmoid} function. Models are trained with mean binary cross-entropy loss between the predicted distributions $\texttt{Sigmoid$(W(H^{CLS}))$}$ and their ground truth labels.

\noindent{\textbf{Self-training.}} We use the finetunned models with labeled data as \textit{teachers} to generate pseudo soft labels on unlabeled data following \citet{Du2020SelftrainingIP}. Pseudo labeled data are combined with original labeled data to trained student models by optimizing objective function in Equation \ref{eq:st_obj}. In the first round, \textit{students} utilize the same pre-trained checkpoints as their teachers and in the following rounds, \textit{students} inherit parameters from \textit{teachers}. 
We set maximum rounds as 3 since we observe that setting a much larger round brings the same results or very marginal gains on both SST-2 and CoNLL 2003. 
\begin{figure*}[t!]
  \centering
  \includegraphics[scale=0.62]{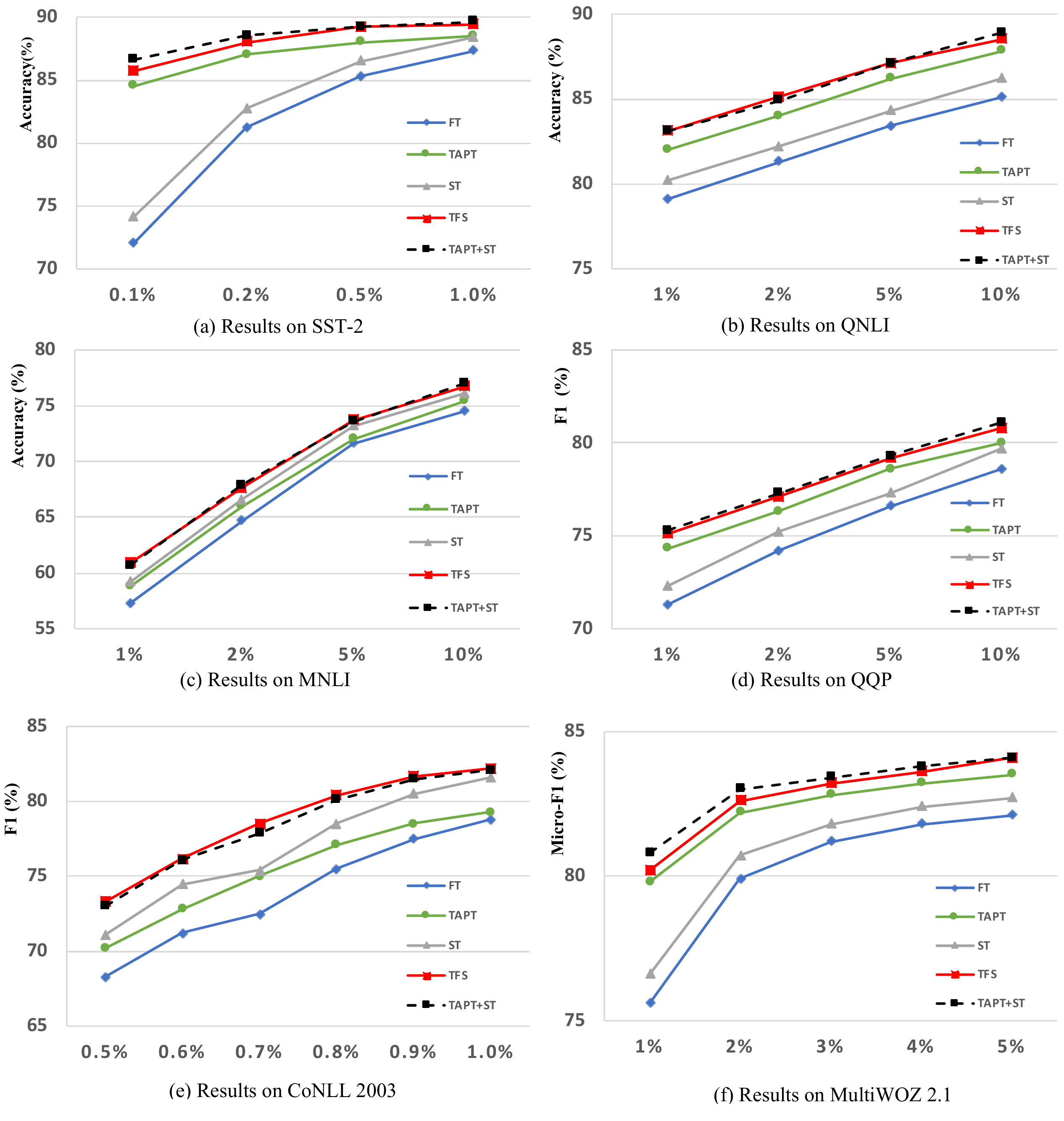}
  \caption{ Results comparison (\%) of FT, TAPT, ST and TFS with different sizes of labeled data on six datasets. \texttt{TAPT+ST} is \textbf{\textit{not}} a method but references to demonstrate additive gains of TFS.}
  \label{fig:ablation}
\end{figure*}

\subsection{Main results} \label{main_result_sec}

In this section, we simulate data scarcity scenarios for these mentioned datasets in Table \ref{statistics} for both $\texttt{BERT}_{\texttt{base}}$ and $\texttt{BERT}_{\texttt{large}}$. Specifically, for each dataset we randomly sample 1\% training data as labeled corpus and left 99\% as unlabeled data. Both labeled and unlabeled corpus is used as the input of TAPT while only unlabeled corpus is used for self-training. For all datasets, we randomly choose three data splits and have three different runs for each of them except $\texttt{BERT}_{\texttt{large}}$ on CoNLL 2003 and MultiWOZ 2.1 to combat their instability by leveraging their results on development sets. In these two datasets, we use ten different runs for each data split on $\texttt{BERT}_{\texttt{large}}$ and report corresponding test set results based on top three runs on development sets. Results are summarized in Table \ref{main_results}.

\noindent \textbf{Comparison between TAPT and ST.}  TAPT and ST in both  $\texttt{BERT}_{\texttt{base}}$ and $\texttt{BERT}_{\texttt{large}}$ models consistently outperform finetuned baseline results across six different datasets, demonstrating their effectiveness as semi-supervised methods on NLU tasks. However, TAPT has inconsistent results compared to ST. TAPT outperforms ST in SST-2, QQP and MultiWOZ 2.1 datasets but underperforms ST or only achieves comparable results for QNLI, MNLI and CoNLL 2003. These results indicate that they learn different representations from unlabeled data since they utilize data from different perspectives.

\noindent \textbf{TFS shows strong \textit{additive} gains over individual TAPT and ST.} On QNLI, TFS on $\texttt{BERT}_{\texttt{base}}$ improves 4\% accuracy over finetuned baselines (FT), equal to the sum of gains from TAPT (+2.9\%) and ST (+1.1\%), and on $\texttt{BERT}_{\texttt{large}}$ improves 1.8\% accuracy, even slightly larger than the sum of gains from TAPT (+0.6\%) and ST (+1.1\%). On MNLI, TFS on $\texttt{BERT}_{\texttt{base}}$ improves 3.6\% accuracy, larger than the sum of gains from TAPT (+1.5\%) and ST (+1.9\%). Similar results on $\texttt{BERT}_{\texttt{base}}$ also hold for CoNLL 2003. For results of other settings, improvements of TFS can also be well approximated by simply adding gains from corresponding TAPT and ST over FT. These consistent and significant results show that TAPT and ST are complementary to each other and TFS can effectively add their gains.

\subsection{Varying size of labeled data}

We have demonstrated the effectiveness of TFS on both $\texttt{BERT}_{\texttt{base}}$  and $\texttt{BERT}_{\texttt{large}}$ in six different datasets with 1\% training data in section \ref{main_result_sec}. 
We further explore different sizes of labeled data on six datasets in Table \ref{statistics} with $\texttt{BERT}_{\texttt{base}}$ model.

Specifically, on relatively simple dataset SST-2, we vary labeled data ratio as $\{0.1\%,0.2\%,0.5\%,1.0\%\}$ and on more difficult QNLI, MNLI and QQP datasets, we vary their labeled ratios as $\{1\%,2\%,5\%,10\%\}$. For CoNLL 2003, we explore labeled data ratio in $\{0.5\%,0.6\%,0.7\%,0.8\%\,0.9\%,1.0\%\}$ and for MultiWOZ 2.1, we set labeled data ratio as $\{1\%,2\%,3\%,4\%,5\% \}$. Following previous settings, for each labeled data ratio among these six datasets, we randomly select 3 data splits and each data split has three different runs. Final average results for each data ratio are reported over these nine runs. To better measure additive property of TFS, we introduce \texttt{TAPT+ST} in our results for references, which directly adds performance gains of TAPT and ST on FT.

Results of six different datasets among different sizes of labeled data are summarized in Figure \ref{fig:ablation}. TAPT outperforms ST in SST-2, QNLI, QQP and MutiWOZ 2.1 datasets in various labeled data setups but underperforms it on MNLI and CoNLL 2003, indicating that TAPT and ST learn different representations from unlabeled data and have their pros and cons. However, TFS consistently and significantly outperforms TAPT and ST in all scenarios among six different datasets and again proves its effectiveness over TAPT and ST alone. For example, in CoNLL 2003 with 0.5\% labeled data, TFS has relative 4.4\% and 3.1\% improvement  over TAPT and ST, respectively. More importantly, TFS overall has very similar results with \texttt{TAPT+ST} in various labeled data size of different datasets, which further strengthens that TFS protocol can yield strong additive gains over
TAPT and ST.

\subsection{Analysis}

Given the promising results in the previous experiments, we aim to answer why TFS outperforms ST consistently and significantly. Indeed, the differences between TFS and ST lie in two aspects: (1) TFS uses initialization from the checkpoint of TAPT rather than original BERT as ST. (2) TFS utilizes pseudo labels generated from \texttt{TAPT finetuned models} while ST uses pseudo labels generated from \texttt{BERT finetuned models} (FT). To further investigate these two perspectives, we design a variant of original ST, \textbf{ST} with \textbf{T}APT \textbf{I}nitilization (STTI), which utilizes pseudo labels generated by BERT finetuned models as ST but is initialized with the same checkpoints from TAPT as TFS during the first round of self-training. The intermediate variant can help us better understand what makes TFS work. We run experiments on SST-2 with 0.1\% and 1.0\% labeled data for $\texttt{BERT}_{\texttt{base}}$ to compare ST, STTI and TFS. The results of STTI are obtained by running over the same three data splits as ST and TFS, and having three different runs for each data split. Results are averaged and summarized in Table \ref{table:st_ablation}. 

\noindent \textbf{Importance of initialization.} Table \ref{table:st_ablation} shows that STTI consistently outperforms ST in both 0.1\% and 1\% labeled setup. Comparing its difference with ST, we can conclude that its improvement over ST comes from its TAPT initialization. Results of MNLI and CoNLL 2003 in Figure \ref{fig:ablation} (c) and (e) also validate the importance of initialization. In these two datasets, although ST can consistently generate more accurate labels than TAPT finetuned models, meaning that it can match TAPT finetuned performance during self-training process, it still underperforms TFS in the end. These results again indicate the importance of initialization. Without TAPT as initialization, even if ST itself can outperform TAPT finetuned models, who are teachers of TFS in self-training process, but still at its end will be left behind of TFS.

\noindent \textbf{Importance of pseudo label correctness.} Table \ref{table:st_ablation} also shows that STTI underperforms TFS in both 0.1\% and 1\% labeled setup although both of them inherit the same parameters from TAPT. These results indicate that beyond initialization, accurate pseudo labels also matter for self-training process. STTI takes pseudo labels generated from BERT finetuned baselines (FT) that have more errors while TFS utilizes more accurate pseudo labels generated from TAPT finetuned models. Suffering from more incorrect pseudo labels in the beginning of self-training process, STTI may converge to a worse local optima than that of TFS. 
This is even more severe when labeled data is 0.1\% and FT is left far behind of TAPT finetuned models, causing that STTI has 10.3\% accuracy gap compared to TFS. These results prove the importance of accurate pseudo labels for self-training. 

Combining these findings, we argue that TFS can outperform ST at least for two reasons: (1) it has a better initialization from TAPT compared to ST from BERT (2) it utilizes more accurate pseudo labels from TAPT finetuned models than ST.

\begin{table}[t!]
\scriptsize
\centering
\begin{tabular}{cccccc}
\toprule
 & FT & TAPT &  ST  &  STTI &  TFS \\
\hline
Init. & BERT & TAPT*  & BERT & TAPT*  & TAPT* \\
\hline
Pseud. & - & - & FT & FT & TAPT \\
\hline
Acc. (0.1\%) & $72.0$ & $84.5$ &$74.1$ & $75.4$ & $ \textbf{85.7}$ \\
\hline
Acc. (1.0\%) & $87.3$ & $88.5$ &$88.4$ & $88.8$ & $ \textbf{89.4}$ \\

\toprule
\end{tabular}
\caption{\label{table:st_ablation} Results comparison of FT, TAPT, ST, STTI and TFS on SST-2 dataset. Rows with \textit{Init.} and \textit{Pseud.} show initialization and pseudo labeler of different models, respectively. Last two rows list accuracy with 0.1\% and 1.0\% labeled training data of these models. * represents models without finetuning on labeled data.
}
\end{table}

\section{Conclusion}

In this paper, we demonstrate that TAPT and ST are complementary for NLU tasks with TFS by following TAPT $\rightarrow$ Finetuning  $\rightarrow$ Self-training process. Our extensive experiments in various semi-supervised setups across six popular datasets show that they are not only complementary but also strongly \textit{additive} with TFS protocol. We further show that TFS outperforms ST through (1) a better initialization from TAPT (2) more accurate predictions from TAPT finetuned models. We hope that TFS could serve as an important semi-supervised baseline for future NLP studies.

\section*{Acknowledgement}
The authors would like to thank the anonymous reviewers for
their thoughtful comments. This research was sponsored in part by Visa Research and the National Science Foundation under Grant No. IIS 1528175. Any opinions, findings, and conclusions or recommendations expressed in this material are those of the authors and do not necessarily reflect the views of the funding agencies.

\bibliography{anthology,custom}
\bibliographystyle{acl_natbib}
\newpage
\appendix

\end{document}